\documentclass[10pt,letterpaper]{article}

\usepackage{arxiv}

\usepackage{comment} % enables the use of multi-line comments (\ifx \fi) 
\usepackage{fullpage} % changes the margin
\usepackage{amsmath,amsfonts,amsthm}
\usepackage{graphicx,float,xcolor,booktabs,bbm,multirow}
\usepackage[pagebackref,breaklinks,colorlinks]{hyperref}
\usepackage{algorithm2e}
\usepackage{graphicx}
\usepackage{subcaption}

\usepackage[capitalize]{cleveref}
\crefname{section}{Sec.}{Secs.}
\Crefname{section}{Section}{Sections}
\Crefname{table}{Table}{Tables}
\crefname{table}{Tab.}{Tabs.}

\makeatletter
\newcommand{\ccol}[2]{%
  \ifmeasuring@
    #2%
  \else
    \makebox[\ifcase\expandafter #1\maxcolumn@widths\fi]{$\displaystyle#2$}%
  \fi
}
\makeatother

\SetKwComment{Comment}{// }{}
\SetKwInput{Input}{Input}
\SetKwInput{Output}{Output}
\RestyleAlgo{ruled}

\begin{document}
\title{Multiple Perturbation Attack: Attack Pixelwise Under Different $\ell_p$-norms For Better Adversarial Performance}

\author{%
    Ngoc N.~Tran \\
    VinAI Research\\
    {\tt\small v.ngoctnq@vinai.io} \\
\and
    Tuan-Anh Bui \\
    Monash University \\
    {\tt\small tuananh.bui@monash.edu} \\
\and
    Dinh Phung \\
    Monash University \\
    {\tt\small dinh.phung@monash.edu}
\and
    Trung Le \\
    Monash University \\
    {\tt\small trunglm@monash.edu} \\
}

\maketitle

%%%%%%%%% ABSTRACT
\begin{abstract}
   Adversarial machine learning has been both a major concern and a hot topic recently, especially with the ubiquitous use of deep neural networks in the current landscape. Adversarial attacks and defenses are usually likened to a cat-and-mouse game in which defenders and attackers evolve over the time. On one hand, the goal is to develop strong and robust deep networks that are resistant to malicious actors. On the other hand, in order to achieve that, we need to devise even stronger adversarial attacks to challenge these defense models. Most of existing attacks employs a single $\ell_p$ distance (commonly, $p\in\{1,2,\infty\}$) to define the concept of closeness and performs steepest gradient ascent w.r.t. this $p$-norm to update all pixels in an adversarial example in the same way. These $\ell_p$ attacks each has its own pros and cons; and there is no single attack that can successfully break through defense models that are robust against multiple $\ell_p$ norms simultaneously. Motivated by these observations, we come up with a natural approach: combining various $\ell_p$ gradient projections on a pixel level to achieve a joint adversarial perturbation. Specifically, we learn how to perturb each pixel to maximize the attack performance, while maintaining the overall visual imperceptibility of adversarial examples. Finally, through various experiments with standardized benchmarks, we show that our method outperforms most current strong attacks across state-of-the-art defense mechanisms, while retaining its ability to remain clean visually. All code used in this work is available at \url{https://anonymous.4open.science/r/mpa}.
\end{abstract}

\section{Introduction}

With the surge of critical real-world machine learning applications \cite{adas,faceid,ssg}, adversarial machine learning are more emphasized than ever. As the stakes are high for these settings, strong and efficient attacks are very welcomed by the malicious actors; and to counteract this, robust defenses are utilized as a prevention measure. Traditionally, these attacks and defenses focus on a single $p$-normed distance - that is, an attacking adversarial example has to be within some $\varepsilon$-ball around the clean example in the $\ell_p$ space to be considered imperceptible to the user, and defenses have to be robust against that threat model. However, each $\ell_p$ attacks have their pros and cons; and $\ell_p$ defenses are not generalizable. On the defending side, the most effective and popular defenses deal with the $\ell_\infty$ threat: while they generalize well to both $\ell_2$ and $\ell_\infty$, they are significantly weak against the $\ell_1$ attack threat (shown in \cref{tab:cifar10,tab:cifar100}). On the attacking side, while $\ell_1$ attacks are very strong against these defenses, they are very visible to the human eye (see \cref{fig:mpa}). Bridging the gap between these different norms, defenses against multiple $\ell_p$ attacks have been developed \cite{Croce2020Provable,msd} with promising results, but attacks have been falling short behind.

In this work, we introduce a novel attack method named Multiple Perturbation Attack (MPA), where we utilize gradients from multiple $p$-norms and combine them to achieve a strong adversary against both state-of-the-art single-norm defenses and multiple-norm robust models. We take inspiration from the multiple-norm defense \cite{msd}, which is essentially standard adversarial training on a multiple-norm adversary. To break the defense, we propose an improvement to said adversary: the previous adversary chooses the best $\ell_p$ adversarial example at every attack iteration, while ours take it one step further and chooses the best $\ell_p$ perturbation \textit{per-pixel}. This simple combination leads to adversarial examples that are much stronger than most single-norm attacks, while still remains visually identical to the original input. For $\ell_\infty$ defenses, with small-sized inputs like CIFAR datasets, our method falls behind only to AutoAttack-$\ell_1$ \cite{autoattack} in reducing robust accuracy, but are significantly less noticeable (see \cref{fig:mpa,tab:metrics-cifar10,tab:metrics-cifar100}). For multiple-norm defenses, and with more realistic datasets like ImageNet, our method are strictly better than any single-norm attacks, while remaining visually innocuous.

\section{Related Works}
\textbf{Adversarial attacks.} Fast Gradient Sign Method \cite{fgsm} was arguably one of the founding works of this whole field, where the author successfully induce model misclassification by adding a clipped gradient to the original input. Then, Projected Gradient Descent \cite{pgd} makes it an iterative algorithm with some additional improvements, and now has become the go-to attack for most applications. Variants of PGD are developed such as APGD \cite{autoattack,apgd_l1}, which are bundled with black-box attacks to create AutoAttack, an attack ensemble that is now commonly used as a benchmarking tool \cite{autoattack}. While the above uses $\ell_p$ distance as the imperceptibility criteria, other works choose different distances (e.g. Wasserstein) to limit visual difference instead \cite{projected_sinkhorn,exact_wasserstein}. Taking the opposite approach, Carlini-Wagner attack \cite{cw} tries to find the successful adversarial example with the minimum distance from the clean one. DeepFool\cite{deepfool} similarly forgoes the imperceptibility requirement and instead only tries to find the closest boundary to cross over.

\textbf{Adversarial defenses.} A natural defense to any attack is to train the model to correctly predict those adversaries - this is called adversarial training (AT), and an early yet popular one is PGD-AT \cite{pgd}. Another successful approach is to smoothen the classification boundary by using regularization to discourage misclassification on similar examples \cite{trades}. Since then, various defenses have popped up, but most of them falls into the trap of utilizing \textit{gradient obfuscation}, which is easily defeated by adaptive attacks \cite{adaptive} and expectation-over-transformation (EoT) \cite{eot}. Making it simpler to detect this pitfall, AutoAttack incorporated EoT in their suite, and created RobustBench \cite{robustbench}, a leaderboard of adversarial defenses to common attack threats. Regarding more novel attack threats, some works aim to simultaneously maximize multiple $\ell_p$-norm robustness to some success \cite{msd,Croce2020Provable}.

\section{Multiple Perturbation Attack}

\begin{figure*}
    \centering
    \includegraphics[width=\linewidth]{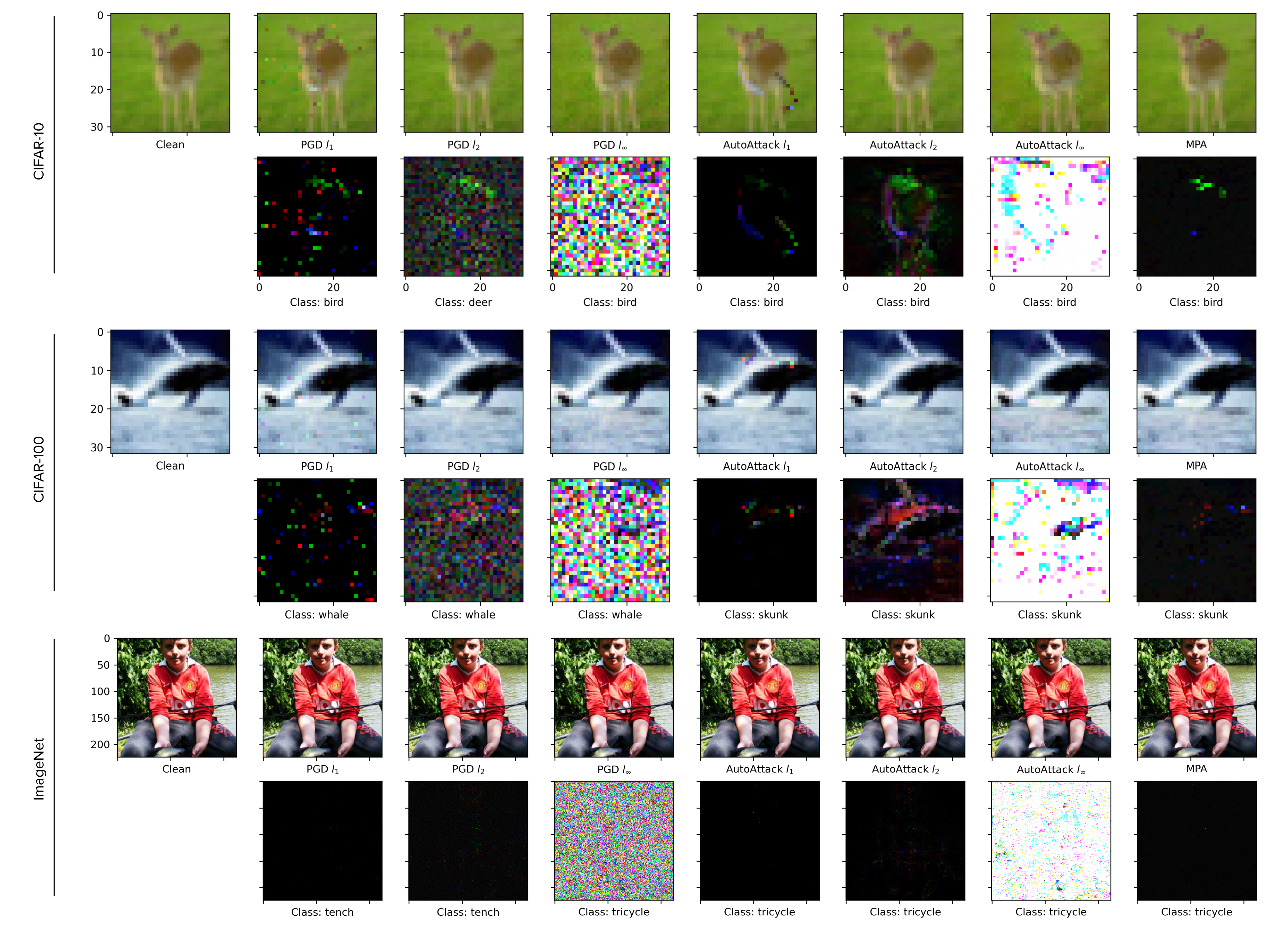}
    \caption{Attack visualization on various datasets. In all cases, our method MPA yields a visually indiscernible result. For CIFAR datasets, while strong, AutoAttack-$\ell_1$ is very obviously altered comparing to the rest. With high-resolution images like ImageNet, sparse but intense perturbations are much less visible.}
    \label{fig:mpa}
\end{figure*}

\subsection{Our Motivations}

It is widely accepted that a good adversarial example has to possess two properties: being able to fool the model (get misclassified), and being identical to a clean example (imperceptible from human perspective). The first criteria can easily be validated by checking whether a model's prediction of the adversarial example is different from which of its corresponding clean one; and the second criteria is commonly verified by using a proper metric to measure distances between adversarial example and its clean example or other metrics to measure their similarity such as Peak Signal-to-Noise Ratio (PSNR), Structural Similarity Index Measure (SSIM), Learned Perceptual Image Patch Similarity (LPIPS) \cite{lpips}, and etc.

Most common $p$-norms have their own pros and cons. $\ell_0$ attacks is very hard to be realized through common optimization methods being a combinatorial problem. $\ell_1$ attacks are very strong against a robust model, but their adversarial examples are very easily detected as they usually have sporadic abnormal pixels across the image (cf. Fig. \ref{fig:mpa}). $\ell_2$ attacks create adversarial examples that are visually identical to their respective clean ones, but the accuracy leaves very much to be desired. As a result, $\ell_\infty$ becomes the de facto threat model for adversarial works.

Maini \textit{et. al} \cite{msd} posed the problem of defending a model against \textit{multiple} $\ell_p$ attacks, and proposed an adversarial training scheme called Multi Steepest Descent (MSD). The difference between MSD and other single-norm defenses is that while the latter trains the model to correctly predict a worst case single-norm adversary, MSD replaces that adversary with another that is good across all $p$-norms. The method of generating MSD adversary also has only one difference comparing to a traditional PGD attack: at each attack iteration, MSD generates an adversarial for each $\ell_p$ threat, and then choose the one with the highest loss to train the model.

While MSD adversaries are good for adversarial training, they fail to significantly reduce the robust accuracy of an existing model. To adapt to the attack settings instead of defense, we propose an improvement: instead of attacking the whole input under a single $p$-norm, we perturb each pixel independently. The intuition is that each pixel may be susceptible to a different $\ell_p$ perturbation, and just forcing the whole image to follow a single $\ell_p$ perturbation might not yield the best adversary. Following the thought, we instead chose the best norm-$p$ perturbation for each pixel independently, while designing a custom projection function that prevents the adversarial example from changing too much.

This novel idea can be intuitively understood as a combination of all $\ell_p$ attacks, whose resulting adversary would likely be strictly ``better'' than each of the independent attacks. For example, if a pixel is predicted to change the input's prediction the most by being perturbed under $\ell_1$, then it should be better than if it were perturbed under $\ell_\infty$; and thus making the combined attack across norms $\mathcal{P}$ better than the pure $\ell_\infty$ attack. This can be observed more closely in the \cref{sec:xp}, where we can see that robust accuracy for our method is strictly lower than those of PGD-$\ell_p$, which are used as our method's building blocks, across all conducted experiments.

\subsection{Our Proposed Attack}
In what follows, we present the technicality of our proposed attack. To simply the presentation, we consider an image $\mathbf{x}\in \mathbb{R}^d$ as a vector in $\mathbb{R}^d$ (i.e., we flatten an image to a vector).

We first initialize the adversarial example for the clean example $\mathbf{x}$ as $\mathbf{x}$ itself, then update the adversarial example for several iterations. Denoting the adversarial example at the start of an iteration as $\mathbf{x}_\mathrm{adv}$, each iteration of our MPA algorithm goes as follows: 

For each norm $p \in \mathcal{P}$ where $\mathcal{P}=\{1,2,\infty\}$ is the set of norms, we compute the steepest ascent direction according to the norm $p$ \cite{msd}. The steepest ascent direction according to the norm $p$ can be understood as the direction that locally maximizes the loss function at $\mathbf{x}_\mathrm{adv}$  w.r.t. its ground-truth label $y$ where the concept of neighborhood is defined via the norm $p$. To accomplish that, we backpropagates through the model computational graph to get the gradient of the loss with respect to the input $\nabla=\partial\mathcal{L}(f(\mathbf{x}),y)/\partial\mathbf{x}$. Then, we get the normalized steepest ascent directions of said gradients across our choice of norms $p\in\mathcal{P}$.

Unlike \cite{msd} where they picked the whole steepest ascent with the highest loss value, we select amongst them independently at each pixel. That is, for each index $i$, the final perturbation at that index belongs to the set $\{\nabla_p^i|p\in\mathcal{P}\}$. The intuition for our approach is that each pixel is vulnerable under a different norm's perturbation, and so we choose one accordingly amongst the norm-$p$ perturbations.

We model this requirement by making the perturbation at index $i$ a weighted sum $\sum_{p\in\mathcal{P}}\mathbf{c}_p^i\nabla_p^i$, where $\mathbf{c}_p^i\in\{0,1\}$ and only one $\mathbf{c}_p^i$ can take on the value $1$. This is an NP-hard combinatorial problem, which we will relax so that it can be solved with traditional gradient descent. To achieve this goal, we opt for using temperature softmax, where we divide the logit by a temperature $\tau>0$ before applying the standard softmax operation. The effect of $\tau$ can be understood as how close the operation reaches hardmax: at $\tau=+\infty$, the result is a uniform distribution with equal values regardless of logit; at $\tau=1$, it's the standard softmax; and as $\tau$ reaches zero, it outputs the one-hot hardmax. 

We then optimize the coefficient tensor $\mathbf{c}$ with SGD to maximize the loss value; and at the end we manually select per-pixel the perturbation with the highest corresponding coefficient. Technically, in Algorithm \ref{alg:mpa}, we update the soft coefficient tensor $\mathbf{c}$ as 

$$
\mathbf{c} = \mathbf{c} + \delta_c \times\\ \dfrac{\partial\mathcal{L}\left(f\left(\mathbf{x}_\mathrm{adv} + \left(\sigma(\frac{\mathbf{c}}{\tau})\odot\begin{bmatrix}\mathbf{\nabla}_{p_1}&\hdots&\mathbf{\nabla}_{p_{|\mathcal{P}|}}\end{bmatrix}^\top\right)\mathbbm{1}^{|\mathcal{P}|}\right), y\right)}{\partial\mathbf{c}},
$$

where $\sigma$ is the softmax function, $\delta_c$ is a learning rate, $\odot$ represents the element-wise product, $\mathcal{P}=\{p_1,\dots,p_{|\mathcal{P}|}\}$ is the set of norms, $\mathbbm{1}^{|\mathcal{P}|}$ is the column vector of all $1$, and the soft coefficient tensor $\mathbf{c} \in \mathbb{R}^{d\times|\mathcal{P}|}$ where $\mathbf{c}^i_p$ specifies the possibility that the pixel $i$ is attacked using the norm $p$, hence $\sigma(\mathbf{c}/\tau)$ represents the probabilities of the pixels under different norm-$p$ attacks.

More specifically, we horizontally stack the normalized norm-$p$ steepest ascent $\mathbf{\nabla}_p$ to obtain a 2D tensor with the dimension $d \times |\mathcal{P}|$, then take element-wise product with $\mathrm{softmax}(\mathbf{c}/\tau)$ and sum up along the columns to yield the mixed-norm perturbation. Finally, we use gradient ascent to update $\mathbf{c}$ to maximize the loss value. This inner optimization process repeats for a predefined number of times. After that inner optimization loop ends, we convert the softmax to a hardmax by just selecting the perturbation with the greatest coefficient at each pixel: at index $i$, the resulting perturbation is then $\nabla^i_{\mathrm{argmax}_p{c^i_p}}$.

An improvement for this mixing is to not reinitialize the coefficients $\mathbf{c}$ after each iteration, since it contains the current information on the importance of each $\nabla_p$ on these pixels, and these perturbations tend to not change too much across iterations. This caching behavior helps the inner optimization loop converges better, resulting in a better attack performance. We examine the effect of this trick with our hyperparameter tuning in \cref{sec:hyperparam}.

The last step of PGD-based attacks is projection, since without which the resulting adversary will be grossly over budget. We simply modify in the same vein as our perturbation mixing method: for any norm $p$, the set of pixels that chose $\nabla_p$ as its perturbation will be projected independently from others, with their separate perturbation budgets under norm-$p$. This custom projection operator is described below in \cref{alg:mpa_combine}.

\begin{algorithm*}

\caption{Combining adversarial perturbations under multiple imperceptibility criteria, with custom mixed projection operation.}
\label{alg:mpa_combine}

\Input{Adversarial image $\mathbf{x}_\mathrm{adv}\in\mathbb{R}^d$, clean image $\mathbf{x}\in\mathbb{R}^d$, set of norms $\mathcal{P}$, mixing weights $\mathbf{c}\in\mathbb{R}^{d\times|\mathcal{P}|}$, maximum budgets $\{\epsilon_p|p\in\mathcal{P}\}$}
\Output{Projected adversarial image $\mathbf{x}_\mathrm{adv} \in \mathbb{R}^d$}

\For{$p\in\mathcal{P}$}{
    \Comment{Get indices where norm-$p$ perturbation will be used}
    $S_p \gets \{i|i\in 1..d,\forall q\in\mathcal{P}:\mathbf{c}_{p}^i\ge\mathbf{c}_{q}^i\}$\;
    \Comment{Add perturbation for each norm}
    $\mathbf{x}_\mathrm{adv}[S_p]\gets\mathbf{x}_\mathrm{adv}[S_p]+\nabla_p[S_p]$\;
    \Comment{Project each sub-image to their respective norm as in \cite{msd}}
    $\mathbf{x}_\mathrm{adv}[S_p]\gets\mathrm{Proj}(\mathbf{x}_\mathrm{adv}[S_p],p,\epsilon_p)$\;
}

\Return $\mathbf{x}_\mathrm{adv}$
\end{algorithm*}

That concludes each iteration of our attack; this will be repeated however many times the user set beforehand. To minimize both computational power and the resulting difference between the adversarial example and its corresponding clean one, we stop our algorithm early at the first iteration it succeeds. A pseudocode of this algorithm is provided in \cref{alg:mpa}. Overall, this attack can be seen as a generalization of PGD, since it directly reduces to the standard single-norm PGD-$\ell_p$ attack if we set $\mathcal{P}=\{p\}$.

\begin{algorithm*}

\caption{Multiple Perturbation Attack (MPA) Algorithm.}
\label{alg:mpa}

\Input{Differentiable classifier function $f$, clean image $\mathbf{x} \in \mathbb{R}^d$, clean label $y$, number of iterations $n$, number of mixing coefficient optimization iterations $n'$, set of norms $\mathcal{P}=\{1,2,\infty\}$, maximum budgets $\{\epsilon_p|p\in\mathcal{P}\}$, step sizes $\{\delta_p|p\in\mathcal{P}\}$, coefficient step size $\delta_c$, softmax temperature $t$}
\Output{Adversarial image $\mathbf{x}_\mathrm{adv} \in \mathbb{R}^d$}

Initialize $\mathbf{x}_\mathrm{adv} \gets \mathbf{x}$\;
Initialize $\mathbf{c} \in\mathbb{R}^{d\times|\mathcal{P}|}$\;
\For{$i = 1..n$}{
  $\mathbb{\nabla} \gets \dfrac{\partial\mathcal{L}\left(f(\mathbf{x}_\mathrm{adv}), y\right)}{\partial\mathbf{x}_\mathrm{adv}}$ \;
  \For{$p\in \mathcal{P}$}{
    \Comment{Follow the steepest ascending direction as described in \cite{msd}}
    $\nabla_p \gets \mathrm{NormalizedSteepestAscent}(\nabla, p, \delta_p)$\;
  }
  \For{$j=1..n'$}{
    \Comment{Use  $\sigma$ = softmax to choose which gradient to be used per pixel}
    % $\mathbf{c} \gets \mathbf{c} + \delta_c\dfrac{\partial\mathcal{L}(f\left(\mathbf{x}_\mathrm{adv} + (\sigma(\mathbf{c}/\tau)\odot\mathrm{hstack}(\{\mathbf{\nabla}_p|p\in\mathcal{P}\}))\mathbbm{1}^{|\mathcal{P}|}\right), y)}{\partial\mathbf{c}}$\;
    $\mathbf{c} \gets \mathbf{c} + \delta_c\dfrac{\partial\mathcal{L}(f\left(\mathbf{x}_\mathrm{adv} + (\sigma(\mathbf{c}/\tau)\odot\begin{bmatrix}\mathbf{\nabla}_{p_1}&\hdots&\mathbf{\nabla}_{p_{|\mathcal{P}|}}\end{bmatrix}^\top)\mathbbm{1}^{|\mathcal{P}|}\right), y)}{\partial\mathbf{c}}$\;
  }
  \Comment{Use hard decision to choose gradient, then custom project}
  $\mathbf{x}_\mathrm{adv} \gets \mathrm{Combine}(\mathbf{x}_\mathrm{adv},\mathbf{x}_0, \mathcal{P}, \mathbf{c},\{\mathbf{\nabla}_p|p\in\mathcal{P}\})$\;
  \If{$f(\mathbf{x}_\mathrm{adv})\neq\mathbf{y}$}{
    \Comment{Stop early if attack succeeds}
    \Return $\mathbf{x}_\mathrm{adv}$
  }
}
\Return $\mathbf{x}_\mathrm{adv}$
\end{algorithm*}

\section{Experiments}
\label{sec:xp}

In this section, we conduct various experiments comparing the performance of our attack algorithm against popular and standardized benchmarking attacks, all of them being evaluated on adversarially-trained models on three datasets including CIFAR-10/100 and ImageNet datasets. We also provide further analyses on the imperceptibility of the adversarial examples and the importance of choosing hyperparameters.

\subsection{Experimental Setup}

\paragraph{Victim models' setting.}
We choose state-of-the-art defense methods on the RobustBench leaderboard \cite{robustbench} as the victim models to evaluate the effective of our proposed attack on the most secure scenario. For the CIFAR-10 dataset (cf. \cref{tab:cifar10}), we choose Rebuffi \textit{et. al}, 2021 \cite{rebuffi21}, Gowal \textit{et. al}, 2021 \cite{gowal21}, Gowal \textit{et. al}, 2020 \cite{gowal20} pre-trained models. 
We also train a multiple-norm robust model (i.e., Maini \textit{et. al}, 2020 \cite{msd} aka MSD) based on their implementation.
% TODO: not really, they train it with 50 attack iter
While the three former models were specific to $\ell_\infty$ attacks, the latter was designed to be robust 
against multiple norm adversarial attack simultaneously, therefore, is expected to be the most 
robust victim model in our setting.
% TODO: also not really, since numbers are down across the board, just more even :)
For the CIFAR-100 dataset (cf \cref{tab:cifar100}), we choose Rebuffi \textit{et. al}, 2021 \cite{rebuffi21}, Gowal \textit{et. al}, 2020 \cite{gowal20} and Debenedetti \textit{et. al}, 2022 \cite{debenedetti22} pretrained models. We also train the MSD model on the dataset. 
For the ImageNet dataset (cf \cref{tab:imagenet}), due to the high computational cost to train a robust model from scratch, we have to choose three pretrained models that are available on the RobustBench including Debenedetti \textit{et. al}, 2022 \cite{debenedetti22}, Salman \textit{et. al}, 2020 \cite{salman20} and Engstorm \textit{et. al}, 2019 \cite{engstrom19}. 

\paragraph{Attack methods' setting.}
Compared with our method are the standard Projected Gradient Descent attack \cite{pgd} and the AutoAttack \cite{autoattack}
adversarial benchmark across different norms. All of our evaluated attacks are set to run for 20 
iterations. Respectively for the norms $\{\ell_1, \ell_2, \ell_\infty\}$, for models trained on the 
CIFAR datasets, the attack budgets are $\{12, 0.5, 0.03\}$, and the per-iteration step sizes 
are $\{0.05, 0.05, 0.003\}$; while for those trained on ImageNet, these numbers are halved accordingly 
to the settings used in RobustBench. Regarding the hyperparameters specific to our method, the inner 
coefficient optimization loop is run for 17 iterations at learning rate $10^{-3}$, the softmax 
temperature is set at $0.01$, and coefficient reuse is enabled. These settings are selected 
after various ablation studies described in \cref{sec:hyperparam}. 
For the CIFAR-10 and CIFAR-100 dataset, we attack on the first 1000 examples similar to \cite{msd} 
while for the ImageNet dataset, we evaluate on the first 5000 examples on the evaluation set as 
the standard setting.
% More specifically, we use the perturbation budget 4/255 and step size 1/255 for attacking the ImageNet dataset.
% NGOC: not really, since the step size is actually set at 1/2/255 (halfed). Since it's too complicated, I didn't give the exact number

\paragraph{Imperceptibility metrics.} To measure the difference of our adversarial example comparing to the clean ones, we opt for using the following metrics: Peak Signal-to-Noise Ratio (PSNR), Structural Similarity Index Measure (SSIM), Learned Perceptual Image Patch Similarity (LPIPS) \cite{lpips}, Wasserstein Distance (WD) \cite{pot}. We do not use RMSE or any $\ell_p$ distances since they use one single norm for the whole input span, while our method processes multiple norms, one for each pixel, at every iteration. One thing to note is that we only measure on the examples that are, 1. originally correctly predicted (else the attack is trivial by adding zero noise), and 2. the attacks succeed on them.

\subsection{Results on ImageNet}

\begin{table*}[!ht]
    \centering
    \caption{Robust accuracy for adversarial-trained models under different attacks on ImageNet (lower is better)}
    \scalebox{0.97}{
    \begin{tabular}{*9c}
        \toprule
        \multirow{2}{*}{Model} & \multirow{2}{*}{Clean} & \multicolumn{3}{c}{Projected Gradient Descent} & \multicolumn{3}{c}{AutoAttack} &\multirow{2}{*}{MPA} \\
        \cmidrule(lr){3-5}\cmidrule(lr){6-8}
        & & PGD-$\ell_1$ & PGD-$\ell_2$ & PGD-$\ell_\infty$ &
        AA-$\ell_1$ & AA-$\ell_2$ & AA-$\ell_\infty$ &\\
        \midrule
        Debenedetti \textit{et. al}, 2022 \cite{debenedetti22} & 79.98\% & 77.96\% & 78.78\% & 69.02\% & 71.32\% & 77.38\% & 55.40\% & \textbf{53.46\%} \\
        Salman \textit{et. al}, 2020 \cite{salman20} & 74.82\% & 69.64\% & 72.68\% & 62.72\% & 50.64\% & 69.66\% & 46.96\% & \textbf{39.36\%} \\
        Engstrom \textit{et. al}, 2019 \cite{engstrom19} & 69.96\% & 65.28\% & 67.98\% & 55.90\% & 44.36\% & 65.00\% & 37.90\% & \textbf{31.70\%} \\
        \bottomrule
    \end{tabular}
    }
    \label{tab:imagenet}
\end{table*}

Tab. \ref{tab:imagenet} reports the evaluation on the ImageNet dataset across various attacks and defenses.
It can be seen that MPA beats every other method by a significant margin: the second best attack performance (i.e. AutoAttack-$\ell_\infty$) is still 
much lower than ours by 2\%, 7.6\% and 6.2\% on the attack success rate on the three defense models, respectively. 
Moreover, since input size is much larger, small perturbations are much less discernible as a result. 
Adversarial examples generated on Engstrom \textit{et. al}'s model and their respective perturbations 
are shown in the visualization provided in \cref{fig:mpa}.

\subsection{Results on CIFAR-10}

\begin{table*}
    \centering
    \caption{Robust accuracy for adversarial-trained models under different attacks on CIFAR-10 (lower is better).}
    \begin{tabular}{*9c}
        \toprule
        \multirow{2}{*}{Model} & \multirow{2}{*}{Clean} & \multicolumn{3}{c}{Projected Gradient Descent} & \multicolumn{3}{c}{AutoAttack} &\multirow{2}{*}{MPA} \\
        \cmidrule(lr){3-5}\cmidrule(lr){6-8}
        & & PGD-$\ell_1$ & PGD-$\ell_2$ & PGD-$\ell_\infty$ &
        AA-$\ell_1$ & AA-$\ell_2$ & AA-$\ell_\infty$ &\\
        \midrule
        Rebuffi \textit{et. al}, 2021 \cite{rebuffi21} & 92.9\% & 41.2\% & 74.9\% & 72.1\% & \textbf{10.7\%} & 68.8\% & 67.3\% & 20.7\% \\
        Gowal \textit{et. al}, 2021 \cite{gowal21} & 89.5\% & 39.8\% & 71.3\% & 70.8\% & \textbf{8.6\%} & 64.1\% & 67.6\% & 21.3\% \\
        Gowal \textit{et. al}, 2020 \cite{gowal20} & 90.7\% & 39.9\% & 73.1\% & 70.7\% & \textbf{7.1\%} & 66.6\% & 67.0\% & 20.7\% \\
        Maini \textit{et. al}, 2020 \cite{msd} & 83.5\% & 62.8\% & 68.4\% & 49.4\% & 49.0\% & 65.9\% & 44.1\% & \textbf{26.0\%} \\
        \bottomrule
    \end{tabular}
    \label{tab:cifar10}
\end{table*}

\cref{tab:cifar10} shows the evaluation on the CIFAR-10 dataset. 
It can be seen that, there is a huge drop on the adversarial robustness of 
the three first defenses method against $\ell_1$ attack. 
This phenomenon can be explained by the fact that all three defense methods were designed for $\ell_\infty$ attack, 
and thus cannot be robust against $\ell_1$ attack. 
In these weak defense models, our MPA attack achieves much higher attack 
success rate than other attacks except AutoAttack with $\ell_1$ with the gap to 
the best of PGD attack (i.e., PGD-$\ell_1$) at around 20\%. For the MSD defense where $\ell_1$ are significantly weaker, our method outperforms every single attack in the list, sitting at 26.0\% robust accuracy. This is 23\% lower than the next highest, which is AutoAttack-$\ell_1$.

It is a worth noting that, while these $\ell_1$ adversaries are powerful, they 
are rather obvious to the naked eye with various oblivious weird pixels scattered across 
the picture, as shown in \cref{fig:mpa}. 
On the other hand, on attacking the MSD model, which is considered as the stronger 
defense model, our MPA attack achieves the best attack performance with the attack success rate 
gap to the second highest (i.e. AA-$\ell_\infty$) is 18\%.

\begin{table}
    \centering
    \caption{Imperceptibility metrics of MPA adversarial examples on CIFAR-10 ($\uparrow$: higher is better, $\downarrow$: lower is better).}
    \begin{tabular}{*5c}
        \toprule
        Attacks & PSNR $\uparrow$ & SSIM $\uparrow$ & LPIPS $\downarrow$ & WD $\downarrow$ \\
        \midrule
        PGD-$\ell_1$ & 31.6006 & 0.9633 & 0.0073 & 0.1650 \\
        PGD-$\ell_2$ & \textbf{40.8943} & 0.9950 & \textbf{0.0012} & \textbf{0.1091} \\
        PGD-$\ell_\infty$ & 31.1556 & 0.9582 & 0.0071 & 0.4867 \\
        \midrule
        AA-$\ell_1$ & 30.9848 & 0.9577 & 0.0117 & 0.1885 \\
        AA-$\ell_2$ & \textbf{40.8943} & \textbf{0.9952} & 0.0013 & 0.1128 \\
        AA-$\ell_\infty$ & 30.7537 & 0.9538 & 0.0079 & 0.5132 \\
        \midrule
        MPA & 32.4387 & 0.9609 & 0.0074 & 0.3801 \\
        \bottomrule
    \end{tabular}
    \label{tab:metrics-cifar10}
\end{table}

In addition to the provided visual in \cref{fig:mpa}, we conduct a further quantitative analysis on the imperceptibility of adversarial attacks in \cref{tab:metrics-cifar10} for a more concrete numerical results.
Amongst the attacks, $\ell_2$ ones give us the cleanest adversaries, but trade off attack efficacy considerably as noted in \cref{tab:cifar10,tab:cifar100}.
Meanwhile, our method's adversarial examples are only behind them on these metrics, while they reduce the model's robust accuracy by a significant margin. Specifically, it can be seen that $\ell_1$ attack has a lower imperceptibility than ours on almost all metrics, which agrees with visual evidence. The only exception to this case is Wasserstein Distance, where $\ell_1$ attacks have a better score than ours: this is due to the fact that this metric requires input normalization beforehand, which spreads out the perturbation mass from being concentrated on a few pixels to the whole image.

To summarize, for weaker models, MPA creates adversarial examples that are much stronger than other imperceptible attacks, while only letting up to $\ell_1$ attacks which both are the natural enemy of $\ell_\infty$ defenses, and give up undetectability for its performance. For stronger multi-norm defenses, MPA gives a much better attack success rate while remaining visually clean.

\subsection{Results on CIFAR-100}

\cref{tab:cifar100} shows the evaluation result matrix across various attacks and defenses. The result this time agrees with our existing findings in every aspect, where our method performs worse only than AutoAttack-$\ell_1$ in $\ell_\infty$ defenses, while achieving the highest robust accuracy in every other scenario. Comparing to the second best attacks (i.e. PGD-$\ell_1$), MPA reduces the robust accuracy to around half of which of PGD-$\ell_1$. Agreeing with our intuitions, the MPA accuracy is always lower than its constituents, showing that MPA successfully select the best $p$-norm perturbation per pixel. On MSD, MPA achieves a 14.0\% robust accuracy, which is 8\% lower than the next lowest one from AutoAttack-$\ell_1$.

\begin{table*}
    \centering
    \caption{Robust accuracy for adversarial-trained models under different attacks on CIFAR-100 (lower is better).}
    \begin{tabular}{*9c}
        \toprule
        \multirow{2}{*}{Model} & \multirow{2}{*}{Clean} & \multicolumn{3}{c}{Projected Gradient Descent} & \multicolumn{3}{c}{AutoAttack} &\multirow{2}{*}{MPA} \\
        \cmidrule(lr){3-5}\cmidrule(lr){6-8}
        & & PGD-$\ell_1$ & PGD-$\ell_2$ & PGD-$\ell_\infty$ &
        AA-$\ell_1$ & AA-$\ell_2$ & AA-$\ell_\infty$ &\\
        \midrule
        Gowal \textit{et. al}, 2020 \cite{gowal20} & 69.3\% & 16.7\% & 45.8\% & 41.1\% & \textbf{4.9\%} & 39.5\% & 35.7\% & 10.3\% \\
        Debenedetti \textit{et. al}, 2022 \cite{debenedetti22} & 70.1\% & 27.8\% & 51.6\% & 39.4\% & \textbf{11.9\%} & 46.0\% & 35.1\% & 14.1\% \\
        Rebuffi \textit{et. al}, 2021 \cite{rebuffi21} & 62.3\% & 20.3\% & 43.7\% & 38.4\% & \textbf{7.3\%} & 39.1\% & 34.3\% & 10.8\% \\
        Maini \textit{et. al}, 2020 \cite{msd} & 56.6\% & 38.9\% & 42.1\% & 25.8\% & 27.4\% & 39.0\% & 22.2\% & \textbf{14.0\%} \\
        \bottomrule
    \end{tabular}
    \label{tab:cifar100}
\end{table*}

We also plot out adversarial examples on Maini \textit{et. al}'s model and their corresponding perturbations in \cref{fig:mpa}. $\ell_1$ attacks' perturbations are still very much noticeable with some out-of-place random-colored pixels, but visually less conspicuous comparing to the CIFAR-10 examples. This is further corroborated with an explicit evaluation of imperceptibility of adversarial attacks in \cref{tab:cifar100}, where the numbers generally agree with which from the CIFAR-10 evaluation.

\begin{table}
    \centering
    \caption{Imperceptibility metrics of MPA adversarial examples on CIFAR-100 ($\uparrow$: higher is better, $\downarrow$: lower is better).}
    \begin{tabular}{*5c}
        \toprule
        Attacks & PSNR $\uparrow$ & SSIM $\uparrow$ & LPIPS $\downarrow$ & WD $\downarrow$ \\
        \midrule
        PGD-$\ell_1$ & 31.8012 & 0.9609 & 0.0077 & 0.1761 \\
        PGD-$\ell_2$ & \textbf{40.8943} & 0.9949 & \textbf{0.0014} & \textbf{0.1293} \\
        PGD-$\ell_\infty$ & 31.1293 & 0.9567 & 0.0091 & 0.5452 \\
        \midrule
        AA-$\ell_1$ & 31.1215 & 0.9577 & 0.0132 & 0.2022 \\
        AA-$\ell_2$ & \textbf{40.8943} & \textbf{0.9953} & 0.0016 & 0.1323 \\
        AA-$\ell_\infty$ & 30.7865 & 0.9525 & 0.0100 & 0.5668 \\
        \midrule
        MPA & 33.1492 & 0.9634 & 0.0075 & 0.3828 \\
        \bottomrule
    \end{tabular}
    \label{tab:metrics-cifar100}
\end{table}

\subsection{Hyperparameter Tuning}
\label{sec:hyperparam}

\begin{figure}
    \centering
    \includegraphics[width=0.6\linewidth]{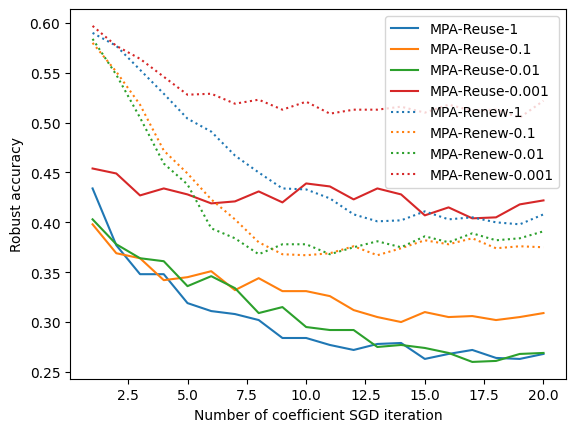}
    \caption{Robust accuracy of models under MPA across different attack hyperparameters.}
    \label{fig:nbiter2}
\end{figure}

Our method has 3 hyperparameters to be selected: whether to reuse the mixing coefficients from previous attack iterations, the softmax temperature, and the number of iterations optimizing the mixing coefficients (i.e. the inner optimization). Running all of these configurations on the MSD defense, we found that reusing coefficients considerably improves our attack performance; especifically at softmax temperature 1 and 0.01, where robust accuracy are very close to the minimum after 15 inner optimization loops. We only experiment with the number of inner optimization iterations up to 20 since at that point, the results plateau across all settings. These results are plotted in \cref{fig:nbiter2}.

\begin{figure}
    \centering
    \includegraphics[width=0.6\linewidth]{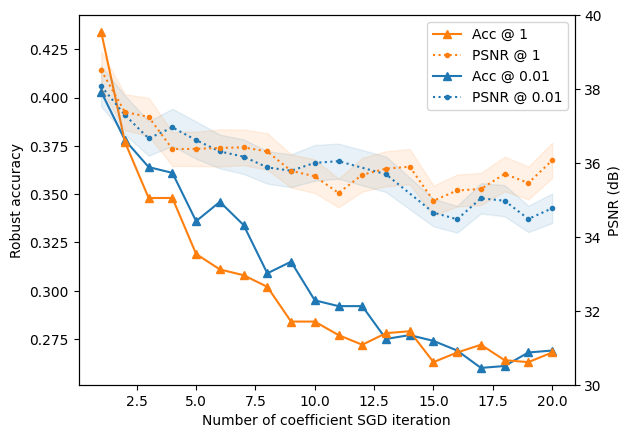}
    \caption{PSNR of MPA adversarial examples across different attack hyperparameters (shades represent 95\% confidence interval).}
    \label{fig:nbiter2_psnr}
\end{figure}

To select between these 2 options as well as the iteration count, we evaluate their ability to stay imperceptible through the PSNR metrics. As we can see in \cref{fig:nbiter2_psnr}, as the inner optimization loop count increases, PSNR generally decreases until hitting a minimum after which it goes back up. At loop count 17, MPA-0.01 gives the lowest robust accuracy, while having a PSNR only slightly lower than which of MPA-1, well within the 95\% confidence interval. Given these results, we recommend using MPA with 17 inner optimization loops, softmax temperature 0.01, and with coefficient reuse.

\section{Discussion}
\paragraph{Limitations.} Since we need to optimize the coefficients, the amount of memory required is tripled (or more specifically $|\mathcal{P}|$-fold). The inner optimization loop has a fixed small learning rate, which are set arbitrarily and perchance work with Kaiming uniform initialization \cite{kaiming} - possibly a different value may fare better with a different coefficient initialization scheme. There is also currently no theory supporting our multiple-norm combination, which we can only offset with extensive empirical experimental results. We invite future works to further analyze our methods.

\paragraph{Social impact.} As with any offensive security, adversarial attacks might be misused in the wrong hands. However, this work is designed to develop the more recent idea that combining multiple $\ell_p$ norms give us better adversarial examples, both in performance and visually. With the back-and-forth nature of adversarial machine learning, stronger attacks should lead to stronger defenses. We hope that this attack will lead to more defenses that create robust models that are resistant to multiple norms simultaneously specifically, and generalize to more novel imperceptibility criteria in general.

\section{Conclusion}
Inspired by a multi-norm defense generating training adversarial examples by selecting the best $\ell_p$ perturbation at every iteration, we propose the Multiple Perturbation Attack (MPA) which takes it a step further and select the $\ell_p$ perturbation \textit{per-pixel} at every iteration. This leads to a significantly higher attack success rate on all state-of-the-art adversarially-trained model comparing to standard attack benchmarks, while remaining visually clean to the human eye. From this work, a natural question arises whether this could improve the standard adversarial training schemes that we already have. Furthermore, we hope that our work would inspire readers to explore new imperceptibility criteria, which might be more practical and effective than the ones commonly used currently.

%% APPENDIX %%
\clearpage

{\Large\bf Supplementary Materials}
\renewcommand{\thesection}{\Alph{section}}
\setcounter{section}{0}

\section{Normalized Steepest Ascent Under Different $p$-norms}

For some norm $p$, a point $\mathbf{z}\in\mathbb{R}^d$, and a radius $\varepsilon>0$, we aim to get the exact solution to the problem:

$$
\mathop{\mathrm{argmax}}_{\Vert\mathbf{v}-\mathbf{z}\Vert_p=\varepsilon}
\mathbf{v}^\top \mathbf{z}.
$$

This gives us the normalized steepest ascent optimization algorithm with step size $\varepsilon$, which has better convergence properties than regular steepest descent (i.e. standard SGD) \cite{nsa}.

\subsection{Normalized Steepest Ascent Under $\ell_\infty$}

The normalized steepest ascent under $\ell_\infty$
norm of a vector $\mathbf{z}$ has the following closed form solution:

$$
\mathop{\mathrm{argmax}}_{\Vert\mathbf{v}-\mathbf{z}\Vert_\infty=\varepsilon}
\mathbf{v}^\top \mathbf{z}=\varepsilon\cdot\mathrm{sign}(\mathbf{z}),
$$

where $\mathrm{sign}$ is the signum function applied elementwise:

$$
\mathrm{sign}(x)=\begin{cases}
1&\text{ if }x>0,\\
-1&\text{ if }x<0,\\
0&\text{ otherwise}.
\end{cases}
$$

\subsection{Normalized Steepest Ascent Under $\ell_2$}

The normalized steepest ascent under $\ell_2$
norm of a vector $\mathbf{z}$ is simply a scaled version of itself:

$$
\mathop{\mathrm{argmax}}_{\Vert\mathbf{v}-\mathbf{z}\Vert_2=\varepsilon}
\mathbf{v}^\top \mathbf{z}=\frac\varepsilon{\Vert\mathbf{z}\Vert_2}\mathbf{z}.
$$

\subsection{Normalized Steepest Ascent Under $\ell_1$}

\paragraph{Canonical version.} The original normalized steepest ascent under $\ell_1$ is

$$
\mathop{\mathrm{argmax}}_{\Vert\mathbf{v}-\mathbf{z}\Vert_1=\varepsilon}
\mathbf{v}^\top \mathbf{z}=\varepsilon\cdot\mathrm{sign}(z_i)\cdot\mathbb{I}_i,
$$

where $i$ is the index that $z_i=\Vert\mathbf{z}\Vert_\infty$, and $\mathbb{I}_i$ is a vector having 0 everywhere except at index $i$. We realistically assume that there's only one $i$ that satisfies the above constraint for simplicity.

\paragraph{Improved version.} Maini \textit{et. al} \cite{msd} improved upon the canonical version to create stronger adversary for both attacking and defending with 2 additions: divide the $\varepsilon$ mass in the top-1 index evenly to top-$k$ indices, and not perturbing indices that would go out of the valid range $[0,1]$ otherwise. We opt to use this version for the experiments in our work.

\section{Projections Under Different $p$-norms}

For some norm $p$, a point $\mathbf{z}\in\mathbb{R}^d$, and a radius $\varepsilon>0$, we aim to get the exact solution to the problem:

$$
\mathop{\mathrm{argmin}}_{\mathbf{v}\in\mathcal{B}_{p,\varepsilon}(\mathbf{z})}
\Vert \mathbf{v} - \mathbf{z}\Vert_2
$$

where

$$
\mathcal{B}_{p,\varepsilon}(\mathbf{z}) = \{
\mathbf{v}\;|\;
\Vert \mathbf{v} - \mathbf{z} \Vert_p\le\varepsilon
\}
$$

is the $\varepsilon$-ball under $p$-norm surrounding the point $\mathbf{z}$.
This guarantees that our perturbation is always within the zero-centered
$\ell_p$-ball of radius $\varepsilon > 0$.

\subsection{Projection Under $\ell_\infty$}

As used in the canonical PGD attack \cite{pgd} under the name \textit{clipping}, the projection operator under $\ell_\infty$
norm has the following closed form solution:

$$
\mathop{\mathrm{argmin}}_{\mathbf{v}\in\mathcal{B}_{\infty,\varepsilon}(\mathbf{z})}
\Vert \mathbf{v} - \mathbf{z}\Vert_2
=\max(\min(\mathbf{z}, \varepsilon), -\varepsilon),
$$

where the $\min$ and $\max$ operations are applied elementwise.

\subsection{Projection Under $\ell_2$}

By definition, the projection under $\ell_2$ is exactly where the $\varepsilon$-ball intersect with $\mathbf{z}$ if they intersect, or $\mathbf{z}$ itself otherwise:

$$
\mathop{\mathrm{argmin}}_{\mathbf{v}\in\mathcal{B}_{2,\varepsilon}(\mathbf{z})}
\Vert \mathbf{v} - \mathbf{z}\Vert_2
=\dfrac{\min(\varepsilon, \Vert\mathbf{z}\Vert_2)}{\Vert\mathbf{z}\Vert_2}\mathbf{z}.
$$

\subsection{Projection Under $\ell_1$}

The projection under $\ell_1$ is the identity function if $\mathbf{z}$ is within the $\varepsilon$-ball. Otherwise, we will apply the adapted simplex projection algorithm from \cite{msd} to solve our problem, restated in \cref{alg:l1_proj}. Derivation of this algorithm can be found in \cite{l1_proj}.

\begin{algorithm}

\caption{Projection onto $\ell_1$-ball}
\label{alg:l1_proj}

\Input{Perturbation $\delta\in\mathbb{R}^d$, ball radius $\varepsilon<\Vert\delta\Vert_1$}
\Output{Projected vector $\mathrm{argmin}_{\Vert\delta'\Vert_1=\varepsilon}\Vert \delta - \delta'\Vert_2$}

Sort $|\delta|$ into $\gamma:\gamma_0\ge\gamma_1\ge\dots\ge\gamma_d$\;
$\rho\gets\max\{j\in[d]:\gamma_j-\frac{1}{j}(\sum_{r=1}^j\gamma_r-\varepsilon)>0\}$\;
$\eta\gets\frac{1}{\rho}(\sum_{i=1}^\rho\gamma_i-\varepsilon)$\;
\For{$i=1..n$}{
    $\gamma'_i\gets\mathrm{sign}(\delta_i)\max(\gamma_i-\eta,0)$\;
}
\Return $\gamma'$
\end{algorithm}

\section{Multiple-Norm Robust Training}

In this section, we summarize the algorithm and settings used to retrain the MSD model \cite{msd} as our multiple-norm robust victim model. This method is a standard adversarial training algorithm solving maximin problem with alternating optimization, where the inner optimization maximizes the loss with respect to the input, and the outer optimization minimizes the loss with respect to the model weights. The specific algorithm for generating MSD adversarial examples is listed in \cref{alg:msd_adv}, and the outer optimization can be done with any optimizer, similar to standard model training.

\begin{algorithm}

\caption{MSD adversary generation}
\label{alg:msd_adv}

\Input{Differentiable classifier function $f$, clean image $\mathbf{x} \in \mathbb{R}^d$, clean label $y$, number of iterations $n$, set of norms $\mathcal{P}=\{1,2,\infty\}$, maximum budgets $\{\epsilon_p|p\in\mathcal{P}\}$, step sizes $\{\delta_p|p\in\mathcal{P}\}$}
\Output{Adversarial image $\mathbf{x}_\mathrm{adv} \in \mathbb{R}^d$}

Initialize $\mathbf{x}_\mathrm{adv} \gets \mathbf{x}$\;
\For{$i = 1..n$}{
  $\mathbb{\nabla} \gets \dfrac{\partial\mathcal{L}\left(f(\mathbf{x}_\mathrm{adv}), y\right)}{\partial\mathbf{x}_\mathrm{adv}}$ \;
  \For{$p\in \mathcal{P}$}{
    \Comment{Norm. steepest ascent}
    $\nabla_p \gets \mathop{\mathrm{argmax}}_{\Vert\mathbf{v}-\mathbf{\nabla}\Vert_p=\delta_p}\mathbf{v}^\top \mathbf{\nabla}$\;
    \Comment{Projection}
    $\mathbf{x}_p\gets\mathop{\mathrm{argmin}}_{\mathbf{v}\in\mathcal{B}_{p,\epsilon_p}(\mathbf{x})}\Vert \mathbf{x}_\mathrm{adv} + \mathbf{\nabla}_p - \mathbf{v}\Vert_2$\;
  }
  \Comment{Select best $\ell_p$ adversary}
  $p'\gets\mathrm{argmax}_{p\in\mathcal{P}}\mathcal{L}(f(\mathbf{x}_p), y)$\;
  \Comment{Clip to valid range}
  $\mathbf{x}_\mathrm{adv}\gets\max(\min(\mathbf{x}_{p'},1),0)$\;
}
\Return $\mathbf{x}_\mathrm{adv}$
\end{algorithm}

As described in the main paper, we set the adversarial example generation hyperparameters same as our attack configuration for a fair comparison. The adversary is generated after 20 iterations, norms $\mathcal{P}=\{1,2,\infty\}$, respective budgets $\{12,0.5,0.03\}$, and respective step sizes $\{0.05, 0.05, 0.003\}$.
\label{sec:atk_params}

\begin{figure}
    \centering
    \includegraphics[width=0.5\linewidth]{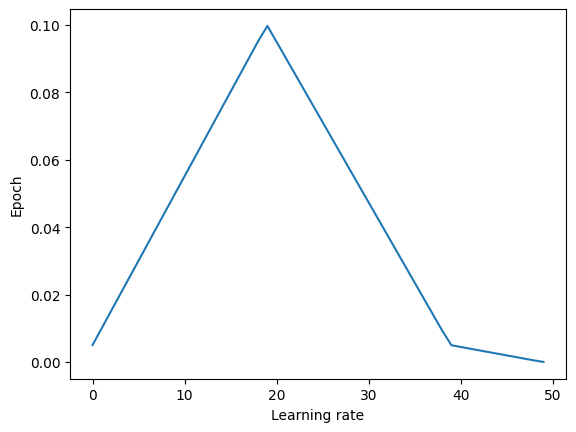}
    \caption{Learning rate schedule for MSD training.}
    \label{fig:lr}
\end{figure}

To optimize the model weights, we use the Stochastic Gradient Descent optimizer for 50 epochs with batch size 128, a custom learning rate scheduler plotted in \cref{fig:lr}, momentum $0.9$, and weight decay $5\times10^{-4}$. This custom learning rate scheduler is taken directly from the MSD code on GitHub.\footnote{\url{https://github.com/locuslab/robust_union/blob/5cd19599b0162beb8803ea6043459fbf7546ad9a/CIFAR10/train.py\#L60}}

\section{Imperceptibility Analysis}

In this section, we present further evidences and experimental results regarding the imperceptibility property of adversarial examples. We first present more results on hyperparameter selection process, then provide more generated adversarial examples and their corresponding differences.
%, and more metrics evaluations on victim models.
For the visualization, we focus on the AutoAttack-$\ell_1$ adversaries unless specified otherwise. Note that we use the recommended \texttt{standard} version of the AutoAttack suite \cite{autoattack}, with budgets listed above in \cref{sec:atk_params} of the Appendix. For the visualization, we attack with batch size 1, and select the first 3 all-success examples in the test set.

\subsection{The Effect Of Hyperparameters On Imperceptibility}

\begin{figure*}
    \begin{subfigure}{0.325\linewidth}
        \centering
        \includegraphics[height=3.6cm]{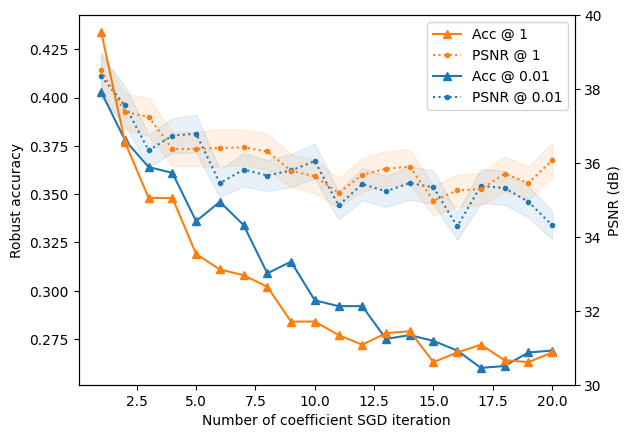}
        \caption{PSNR $\uparrow$}
    \end{subfigure}%
    ~ 
    \begin{subfigure}{0.325\linewidth}
        \centering
        \includegraphics[width=\linewidth]{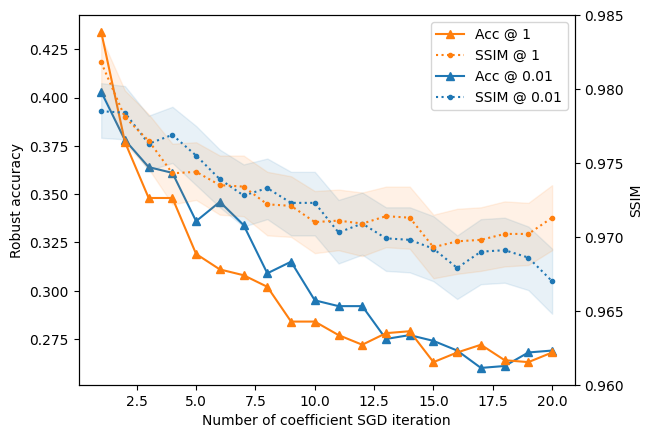}
        \caption{SSIM $\uparrow$}
    \end{subfigure}%
    ~ 
    \begin{subfigure}{0.325\linewidth}
        \centering
        \includegraphics[width=\linewidth]{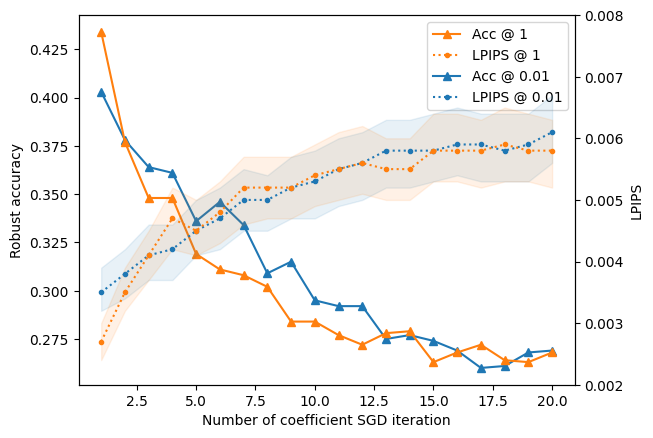}
        \caption{LPIPS $\downarrow$}
    \end{subfigure}
    \caption{Imperceptibility metrics across different top-2 hyperparameter choices ($\uparrow$: higher is better, $\downarrow$: lower is better)}
    \label{fig:hyperparams}
\end{figure*}

After seeing that every other choice of hyperparameters perform significantly inferior to the top 2 candidates, we measure how their resulting adversarial examples differ from the original through metrics. The trend concurs across all PSNR, SSIM, and LPIPS: firstly, the perturbation noise gets more noticeable as the number of inner optimization loop count increases. Also, the optimal choice stays MPA-0.01 at 17 inner loop count: while robust accuracy is reduced to a minimum, its metrics values are all well within the 95\% confidence interval of which of MPA-1. These plots are presented in \cref{fig:hyperparams}.

\subsection{Results on CIFAR-10}

\begin{figure*}
    \centering
    \includegraphics[width=\linewidth]{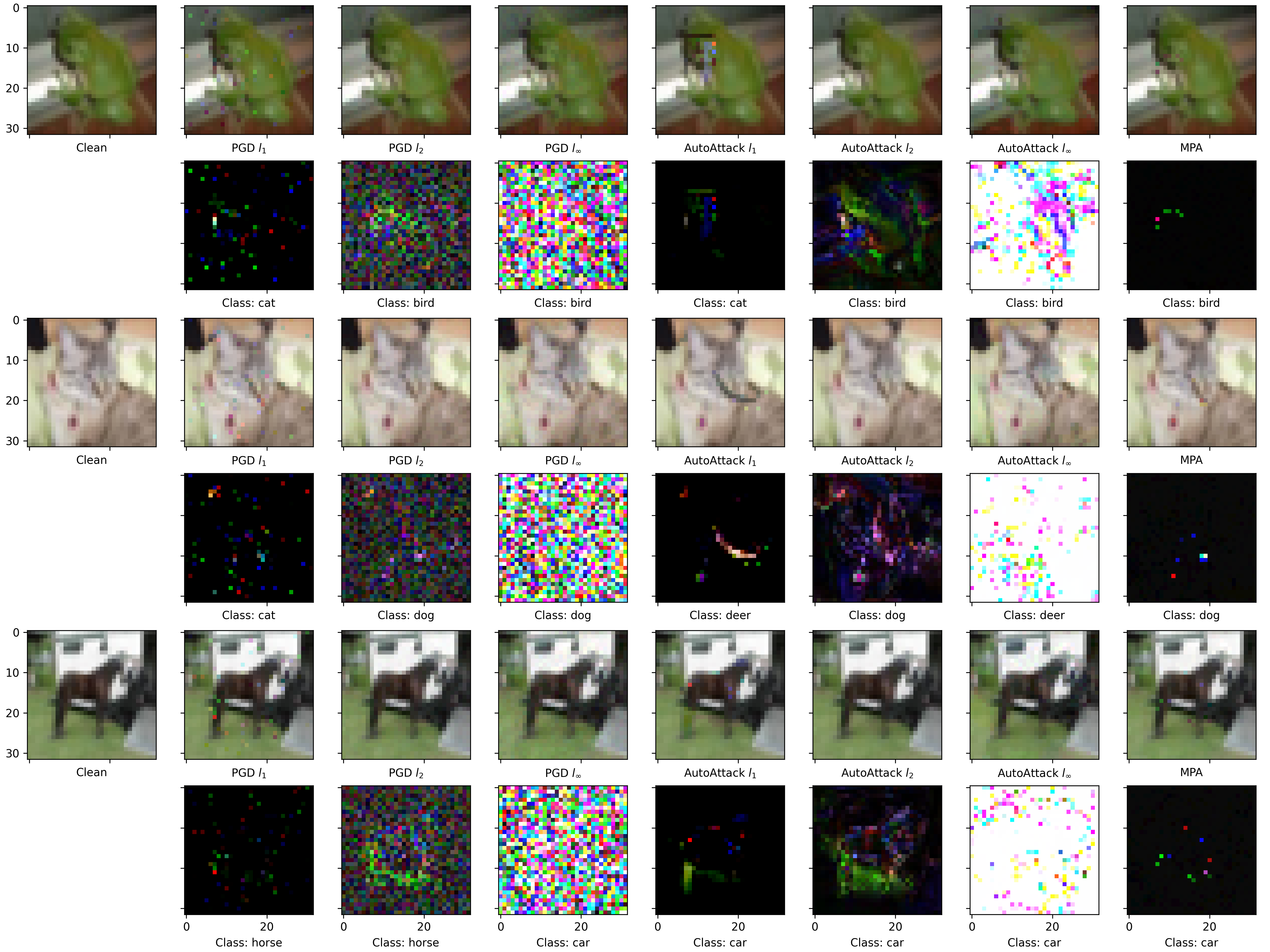}
    \caption{Successful attacks and their respective perturbations on CIFAR-10 \cite{msd}. $\ell_1$ attacks are very visually apparent.}
    \label{fig:cifar10_2}
\end{figure*}

We provide some more visual examples for successful attacks on the CIFAR-10 dataset in \cref{fig:cifar10_2}. While it remains true that $\ell_1$ attacks are pretty obvious visually and our method creates clean adversarial examples, the details of the AA-$\ell_1$ ones are rather interesting. The first image, which is of a frog, is rather standard with random off-color pixels scattered in some image region. The second image of a cat however, besides 3 visibly strange pixels, has a clean ring edited around the cat's nose. The third one of a horse, from first look has only one strange red pixel on his behind; but if we pay attention with the help of the perturbation heatmap below, itss back legs are faded out neatly. These results show that AA-$\ell_1$ examples are obvious due to not only random pixels standing out, but sometimes also semantically unusual visual cues.

\begin{table}
    \centering
    \caption{Imperceptibility metrics of adversarial examples on \cite{rebuffi21} for CIFAR-10 ($\uparrow$: higher is better, $\downarrow$: lower is better).}
    \begin{tabular}{*5c}
        \toprule
        Attacks & PSNR $\uparrow$ & SSIM $\uparrow$ & LPIPS $\downarrow$ & WD $\downarrow$ \\
        \midrule
        PGD-$\ell_1$ & 30.1971 & 0.9438 & 0.0080 & 0.1408 \\
        PGD-$\ell_2$ & \textbf{40.8943} & 0.9930 & \textbf{0.0010} & 0.0625 \\
        PGD-$\ell_\infty$ & 31.2749 & 0.9562 & 0.0076 & 0.5644 \\
        \midrule
        AA-$\ell_1$ & 28.8350 & 0.9327 & 0.0096 & 0.1663 \\
        AA-$\ell_2$ & \textbf{40.8943} & \textbf{0.9932} & \textbf{0.0010} & \textbf{0.0558} \\
        AA-$\ell_\infty$ & 30.9507 & 0.9533 & 0.0085 & 0.5808 \\
        \midrule
        MPA & 31.8752 & 0.9562 & 0.0069 & 0.3840 \\
        \bottomrule
    \end{tabular}
    \label{tab:cifar10-rebuffi21}
\end{table}

We also run metrics evaluation on \cite{rebuffi21} for CIFAR-10 for a second take and found out that the numbers do not differ much from our results with \cite{msd}, where MPA is better than every other non-$\ell_2$ attacks for all metrics but Wasserstein; and for Wasserstein $\ell_1$ adversaries yield lower numbers than MPA due to mass normalization. The specific numbers are listed in \cref{tab:cifar10-rebuffi21}.

\subsection{Results on CIFAR-100}

\begin{figure*}
    \centering
    \includegraphics[width=\linewidth]{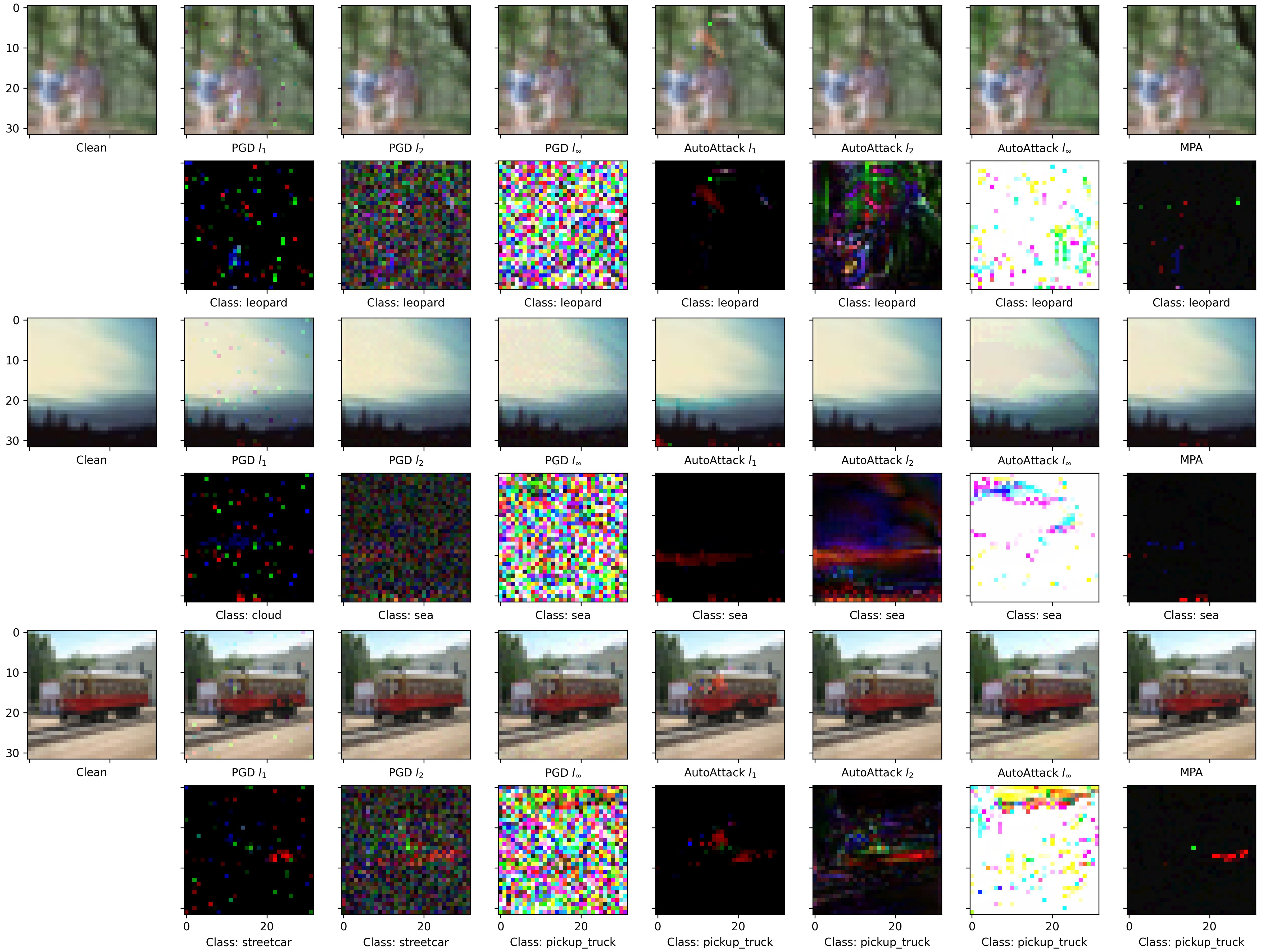}
    \caption{Successful attacks and their respective perturbations on CIFAR-100 \cite{msd}. $\ell_1$ attacks are still detectable with human eye.}
    \label{fig:cifar100_2}
\end{figure*}

While $\ell_1$ attacks still contains visible artifacts and MPA still looks generally clean, the case for CIFAR-100 is slightly different (cf. \cref{fig:cifar100_2}). For the first image of two people, there's a really bright neon-green pixel, and a light pink stroke in the image. For the second image of the sky, there's also a blue halo at the bottom of the horizon, which actually looked completely natural and makes the image better. However, there are still red pixels at the bottom of the adversarial examples, which to be fair MPA also has but less flagrant. For the third picture of a tram, AA-$\ell_1$ spreads vibrant colors on the vehicle's head and thus makes it really visible; while with MPA the adversarial example only adds more of the same color onto the tram's body, blending it in.

\subsection{Results on ImageNet}

\begin{table}
    \centering
    \caption{Imperceptibility metrics of adversarial examples on \cite{engstrom19} for ImageNet ($\uparrow$: higher is better, $\downarrow$: lower is better).}
    \begin{tabular}{*5c}
        \toprule
        Attacks & PSNR $\uparrow$ & SSIM $\uparrow$ & LPIPS $\downarrow$ \\
        \midrule
        PGD-$\ell_1$ & 50.2323 & 0.9994 & 0.0009 \\
        PGD-$\ell_2$ & 63.7140 & 0.9999 & \textbf{0.0001} \\
        PGD-$\ell_\infty$ & 37.3261 & 0.9707 & 0.0003 \\
        \midrule
        AA-$\ell_1$ & 47.2589 & 0.9988 & 0.0026 \\
        AA-$\ell_2$ & \textbf{63.7141} & \textbf{1.0000} & \textbf{0.0001} \\
        AA-$\ell_\infty$ & 36.7596 & 0.9648 & 0.0410 \\
        \midrule
        MPA & 36.9276 & 0.9617 & 0.0005 \\
        \bottomrule
    \end{tabular}
    \label{tab:metrics-imagenet}
\end{table}

\begin{figure*}
    \centering
    \includegraphics[width=\linewidth]{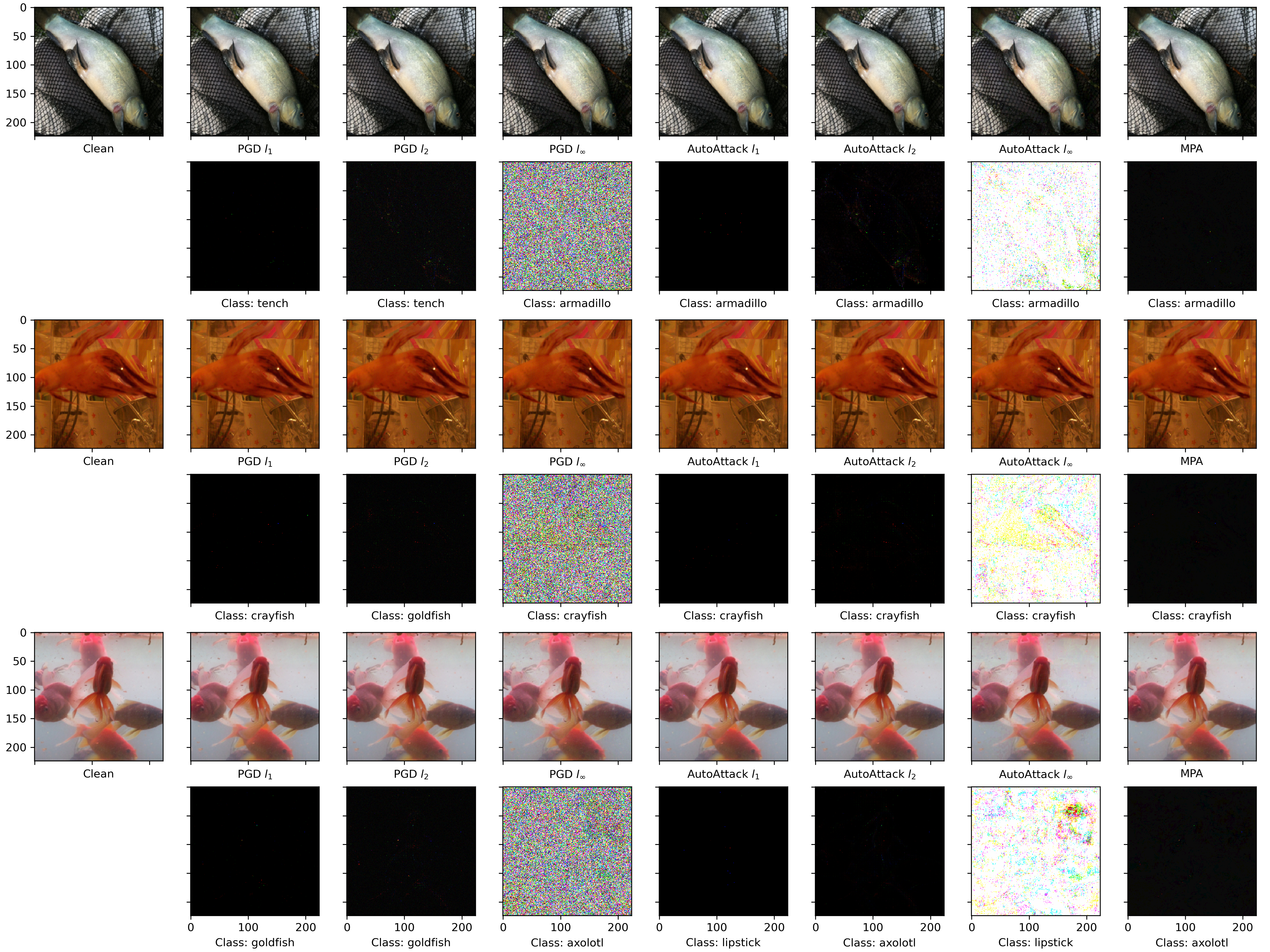}
    \caption{Successful attacks and their respective perturbations on ImageNet \cite{engstrom19}. For large-sized inputs like ImageNet, any reasonable perturbation is practically invisible.}
    \label{fig:imagenet2}
\end{figure*}

For completeness' sake, in addition to the imperceptibility metrics provided in the main paper, we run the same metrics against various attacks on \cite{engstrom19} for ImageNet, excluding Wasserstein Distance \cite{pot} since the computation exceeds the maximum number of iterations for a large-sized input. The numerical results are listed in \cref{tab:metrics-imagenet}. Similarly to the example visualization provided in \cref{fig:imagenet2}, the numbers do not say much either. Specifically, for PSNR, let us recall the calculation formula as:

$$
\mathrm{PSNR}(\mathbf{x}, \mathbf{\delta}) =20\log_{10}\Vert\mathbf{x}\Vert_\infty - 10\log_{10} \Vert\delta\Vert_2^2 + 10\log_{10}\dim\mathbf{x}
$$

where $\mathbf{x}$ is the original image, $\mathbf{\delta}$ is the final perturbation, and the last summand is a constant term only depending on the size of the image. As $\ell_\infty$ perturbs every pixel by a small magnitude, they add up quickly. The root problem is rather because PSNR is a basically $\ell_2$: the first norm term is usually 1 since most pictures contain a white pixel, and the second term is solely dependent on the maximum $\ell_2$ budget. This is very relevant to metrics alignment \cite{adv_bugs}, and might explain the reason why $\ell_\infty$ defenses are weak against $\ell_1$ attacks.

For the other two metrics, evaluation yields near-optimal results that it does not imply much. Specifically, SSIM numbers are all close to the largest possible value of 1; and LPIPS numbers are all close to the smallest possible value of 0. We provide some more visual examples for successful attacks on the ImageNet dataset in \cref{fig:imagenet2}. As we can see, the standard attack settings yield perturbations practically invisible to the human eye.

\clearpage

\bibliographystyle{plain}
\bibliography{ref}

\end{document}